\documentclass[preprint,12pt]{elsarticle}

\usepackage{amssymb}

\usepackage{amssymb}
\usepackage{microtype}
\usepackage{multirow}
\usepackage{booktabs}
\usepackage{float}
\usepackage{amsmath}
\usepackage{lipsum}
\usepackage{multirow}
\usepackage{algorithmic}
\usepackage{algorithm}

\usepackage[algo2e,ruled,vlined]{algorithm2e}

\usepackage{makecell}
\usepackage[caption=false,font=tiny,labelfont=rm,textfont=rm]{subfig}
\bibpunct{[}{]}{,}{a}{}{;}

\begin{document}

\begin{frontmatter}

\title{Query-Efficient Hard-Label Black-Box Attack against Vision Transformers}


\author[inst1]{Chao Zhou}
\ead{zhouchaodzkd@e.gzhu.edu.cn}
\author[inst1]{Xiaowen Shi}
\ead{shixiaowen@e.gzhu.edu.cn}


\author[inst1]{Yuan-Gen Wang}
\ead{wangyg@gzhu.edu.cn}

\affiliation[inst1]{organization={School of Computer Science and Cyber Engineering},
            addressline={Guangzhou University}, 
            city={Guangzhou 510006},
            country={China}
            }


\begin{abstract}
Recent studies have revealed that vision transformers (ViTs) face similar security risks from adversarial attacks as deep convolutional neural networks (CNNs). However, directly applying attack methodology on CNNs to ViTs has been demonstrated to be ineffective since the ViTs typically work on patch-wise encoding. This article explores the vulnerability of ViTs against adversarial attacks under a black-box scenario, and proposes a novel query-efficient hard-label adversarial attack method called AdvViT. Specifically, considering that ViTs are highly sensitive to patch modification, we propose to optimize the adversarial perturbation on the individual patches. To reduce the dimension of perturbation search space, we modify only a handful of low-frequency components of each patch. Moreover, we design a weight mask matrix for all patches to further optimize the perturbation on different regions of a whole image. We test six mainstream ViT backbones on the ImageNet-1k dataset. Experimental results show that compared with the state-of-the-art attacks on CNNs, our AdvViT achieves much lower $L_2$-norm distortion under the same query budget, sufficiently validating the vulnerability of ViTs against adversarial attacks.
\end{abstract}




\end{frontmatter}


\section{Introduction}
\label{sec:intro}
The recent decade has witnessed that deep convolutional neural networks (CNNs) may make the wrong classification decision on the input images added with almost imperceptible perturbations to the naked eye. These perturbed images are called adversarial examples, whose classes can be easily recognized by humans~\citep{szegedy2014intriguing}. Nowadays, vision transformers (ViTs) have sparked a new wave in network architecture design thanks to their record-breaking performance in various vision tasks. As reported, ViTs may raise the same security issues as CNNs~\citep{KOTYAN2024127000},~\citep{bhojanapalli2021},~\citep{fu2022patch}. However, recent studies on the robustness of ViTs have discovered that ViTs are more robust to adversarial attacks than CNNs~\citep{bhojanapalli2021},~\citep{shao2022adversarial},~\citep{mahmood2021},~\citep{wei2022towards},~\citep{mahmood2021robustness}. Such better robustness is interpreted as that the ViTs focus on capturing global interactions between different patches. By now, these results have been achieved with the attack methods developed for CNNs, and few attack methods are designed for ViTs in hard-label black-box scenarios.

As far as we know, Shi~\textit{et al.}~\citep{shi2022decision} first attempted to attack the ViTs under a black-box setting, and developed a from-coarse-to-fine patch size noise removal method (PAR). PAR divides the initially perturbed images into multiple large-size patches and uses two masks to record the noise sensitivity and noise amplitude of each patch. It then removes the noise from large-size to small-size patches gradually. However, PAR allows the removal of noise within the patches in a binary form, i.e., only two states of preservation or erasure. This limits the optimization for the noise amplitude within the patches, leading to visually inferior adversarial examples. In addition, PAR does not consider the adaption to the patch-based encoding specifics within the ViTs. Thus, PAR can only be used as a noise initialization backbone for other decision-based adversarial attacks. On the one hand, ViTs divide the images into a set of patches and encode them into the corresponding vectors with the same length. It can help to learn the long-range dependencies between different regions. However, this strategy makes the ViT models highly sensitive to modification of a single encoding patch~\citep{fu2022patch}. On the other hand, the major semantics of natural images are often determined by the low-frequency component, while the high-frequency signals in images are usually related to noise~\citep{guo2020low}. Therefore, perturbing the low-frequency components of images (especially for individual patches) will pose great challenges to the robustness of the ViT-based classifiers. Compared to the low-frequency regions, the human visual system is less sensitive to the changes in the high-frequency regions of the images. The adversarial examples can possess a better visual hiding effect when properly concentrating their perturbation on these patches with richer high-frequency information.

\begin{figure*}[tp]
	\centering	\includegraphics[width=1\textwidth]{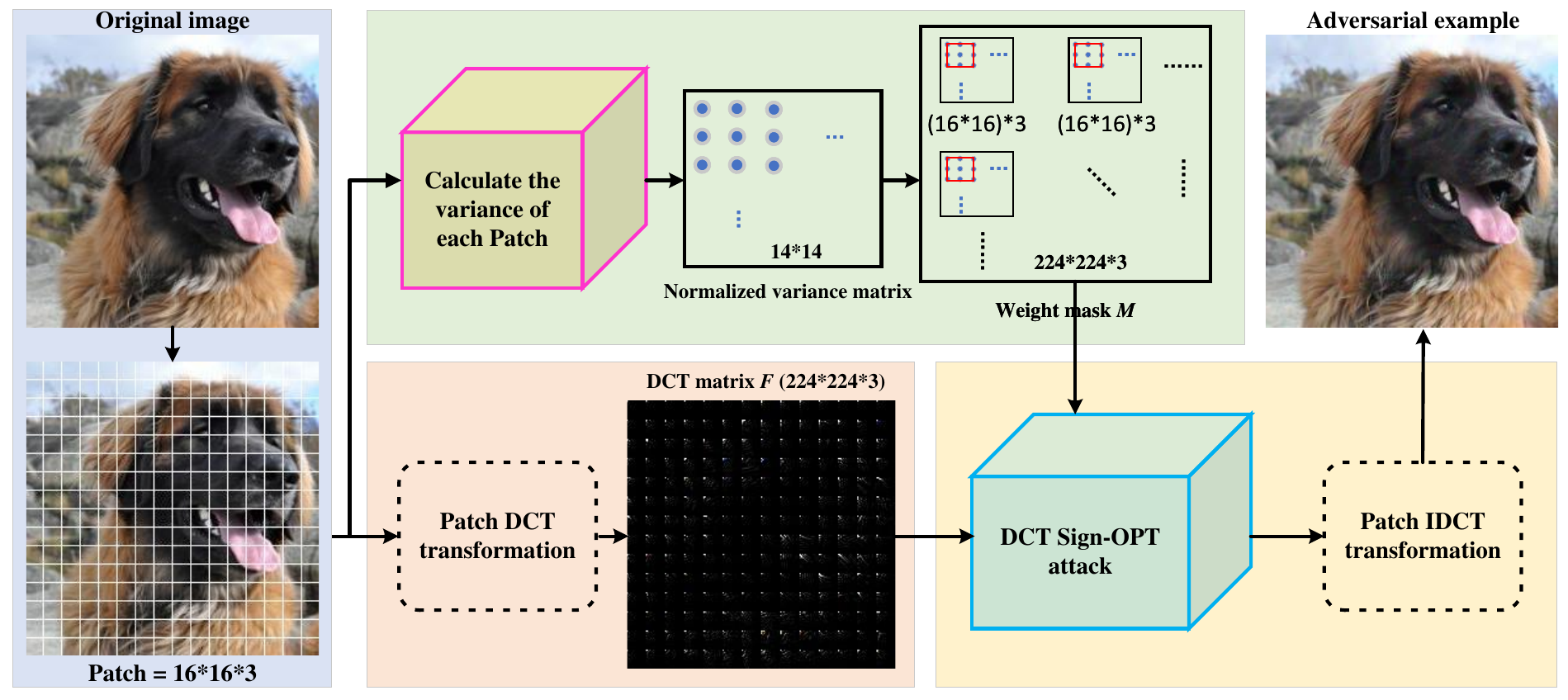}
	\caption{Overview of the proposed AdvViT method.}
	 \vspace{-0.5cm}
   	\label{fig2}
\end{figure*} 

Inspired by the above facts, in this paper, we propose a novel query-efficient black-box adversarial attack against ViTs called AdvViT, which includes three core designs.  
Firstly, we perform a block DCT (discrete cosine transformation) on all channels of every patch ($16 \times 16 \times 3$ size). Then, we concentrate the adversarial attack on individual patches, making it easy to capture the vulnerability of such patch-based ViTs. Secondly, we only attack the low-frequency components of individual patches by designing a $0/1$ mask matrix with only $3 \times 3$ positions of $1$ in the upper left corner and $0$ in all other positions. 
Through this strategy, we can efficiently search adversarial examples in a much lower-dimensional space. Thirdly, we calculate the variance of pixel values for every patch and normalize them by the maximum normalization algorithm. Based on this, we can obtain a weight mask matrix $M$ by multiplying the values of each patch position in the 0/1 matrix by the normalized variance value corresponding to that patch. Matrix $M$ helps to highlight the importance of each patch when generating perturbation, greatly improving the visual hiding effects. Based on these three core designs, our AdvViT is query-efficient and the generated adversarial examples are high-quality. The major contributions of this paper can be summarized as follows.
\begin{itemize}
	\item We investigate the intrinsic specifics within ViTs, and propose a ViTs-adaption adversarial attack method (AdvViT). AdvViT works in the low-frequency component of individual patches. To our knowledge, this work is the first to consider the patch-based nature of ViTs, and launch an effective adversarial attack against ViTs in hard-label black-box settings.
 
	\item We design a weight mask matrix based on the variance of the patch. The weight mask matrix plays a role in concentrating the generated perturbations to the high-frequency regions of the whole image, leading to outstanding visual hiding effects.
 	
	\item Extensive experiments on various ViT backbones demonstrate that, with the same query budget, the adversarial examples generated by our AdvViT method have lower $L_2$-norm distortion, and higher SSIM and PSNR values than that of the state-of-the-art.
\end{itemize}

\section{Related Works}
\subsection{Vision Transformers}
Self-attention-based Transformer was first proposed for natural language processing~\citep{vaswani2017attention}~\citep{GONZALEZ202158}, which can globally model each element in a sequence and establish connections between all elements. Subsequently, Dosovitskiy \textit{et al.}~\citep{dosovitskiy2020image} developed vision transformers (ViTs) and applied them to image classification tasks~\citep{li2021efficient},~\citep{WANG2024127216}~\citep{beal2020toward},~\citep{zheng2021rethinking}. To improve the training efficiency, Touvron \textit{et al.}~\citep{touvron2021training} proposed an improved ViT called Data efficient Image Transformers (DeiT). DeiT exploits knowledge distillation~\citep{hinton2015distilling} and data augmentation to improve ViTs when training with less data. To overcome the singulation of tokenization, Yuan \textit{et al.}~\citep{yuan2021tokens} proposed a Tokens-To-Token ViT (T2T-ViT), which constructs tokens by aggregating neighboring tokens into one token, i.e., tokens-to-token. Swin Transformers ~\citep{liu2021swin},~\citep{liu2022video},~\citep{dong2022cswin} focused on large-size images and effectively solved their feature extraction problems by dividing images into different levels and introducing a local self-attention module. Additionally, researchers have hybridized ViTs with traditional CNNs ~\citep{xiao2021early},~\citep{wu2021cvt},~\citep{graham2021levit}, allowing these models to benefit from both the global feature of ViTs and the local feature of CNNs.
To improve the accuracy and robustness of ViT, Shashank\textit{et al.}~\citep{KOTYAN2024127000} proposed an augmentation strategy named ‘Dynamic Scanning Augmentation’ which can input sequences dynamically and focus on different patches adaptively.

\subsection{Adversarial Attacks}
The concept of adversarial examples was first proposed by Szegedy \textit{et al.} \citep{szegedy2014intriguing}. According to whether the internal parameters and architecture of the model are accessible, the adversarial attacks are mainly divided into white-box attacks and black-box attacks. Thanks to the ease of optimization, lots of works have been proposed in the white-box setting, such as FGSM~\citep{GoodfellowSS14}, I-FGSM~\citep{kurakin2018adversarial}, PGD~\citep{madry2017towards}, JSMA~\citep{papernot2016limitations}, BPDA~\citep{athalye2018obfuscated}, C\&W attack~\citep{carlini2017towards}, L-BFGS~\citep{szegedy2014intriguing}, ANTs~\citep{baluja2017adversarial}, and DeepFool~\citep{moosavi2016deepfool}. The white-box attacks can achieve a very high success rate, however, are generally impractical due to the unavailability of the internal structure of the model. Instead, the black-box attacks (including soft-label and hard-label) were widely developed in practice. For the soft-label setting, attackers can obtain the output probabilities of all the categories, such as ZOO attack~\citep{chen2017zoo}, NES attack~\citep{ilyas2018black}, AutoZOOM attack~\citep{tu2019autozoom}, SimBA attack~\citep{guo2019simple}, LeBA attack~\citep{yang2020learning}, Sparse-RS attack~\citep{croce2022sparse}. Unlike soft-label attacks, hard-label attacks can access only the top-1 prediction class. Hence, the hard-label setting is more secure due to the least information leakage, but is much harder to attack. Brendel\textit{et al.}~\citep{brendel2018decision} proposed the boundary attack, which is 
the first hard-label attack. Based on the boundary attack, Cheng \textit{et al.}~\citep{cheng2019query} treated the hard-label attack as an optimization problem and proposed an Opt attack. Cheng \textit{et al.}~\citep{ChengSCC0H20} designed Sign-OPT attack based on SignSGD algorithm~\citep{liu2018signsgd}. Afterward, Ran \textit{et al.}~\citep{ran2022sign} proposed the Sign-OPT+ method and further improved the query efficiency of Sign-OPT.

\subsection{Robustness of Vision Transformers}
Due to the increasing deployment of ViTs in real-world computer vision tasks, the robustness of the ViTs has attracted great attention. Existing adversarial attacks against ViTs are mostly based on either the white-box setting \citep{bhojanapalli2021},~\citep{shao2022adversarial},~\citep{mahmood2021} or transfer-based black-box setting~\citep{wei2022towards},~\citep{wang2022generating}. As we discussed above, the former is impractical, however, the latter consumes an extremely large query budget. Recent studies have revealed that ViTs are more robust to adversarial attacks than CNNs~\citep{bhojanapalli2021},~\citep{shao2022adversarial},~\citep{mahmood2021},~\citep{wei2022towards},~\citep{mahmood2021robustness}. Soon afterwards, Fu \textit{et al.}~\citep{fu2022patch} demonstrated for the first time that ViTs are not necessarily more robust than CNNs and are highly sensitive to modifications on individual patches. For the hard-label setting, Shi~\textit{et al.}~\citep{shi2022decision} proposed a coarse-to-fine method (PAR) for the patch sizes to remove the noise within the patches.
However, PAR concentrates the perturbations to a small number of patches, resulting in poor visual effects and low query efficiency.

\section{Proposed Method}
Denote $f(\cdot):X^{W\times H \times C}\to y_{K}$ as a black-box classifier where 
$X$ is the model input. $W$, $H$, and $C$ represent the input width, height, and channel, respectively. $y_K$ denotes the classification result with $K$ categories. For a test sample $x_0\in R^{W\times H \times C}$ with ground truth $y_0$, if $y_0 = f(x_0)$, it can be said that the image $x_0$ is classified correctly. For a hard-label black-box model, we consider the process of finding adversarial examples as an optimization problem. The objective function can be written as: 
\begin{equation}
	\begin{split}
		\underset{\hat{x}\in R^{W\times H \times C}}{\textrm{arg min}} 
		\parallel \hat{x} - x_0\parallel_2,\,\, \textrm{s.t.} \,\,  f(\hat{x}) \neq f(x_0),\\
	\end{split}
\end{equation}
where $\parallel \cdot \parallel_2$ represents the $L_2$ norm operation used to measure the distortion.

Unlike previous attack methods designed for CNN models, our method considers the specifics of ViTs and divides images into individual blocks as attack units. Then, we concentrate the adversarial perturbation on the low-frequency component of each patch, which makes it easier to change the semantics of images in the feature space. Finally, we control the perturbation ratio by designing a weight matrix for each patch. Figure~\ref{fig2} shows the overall framework of our method, which includes the patch-adaption perturbation, the weight mask matrix, and the low-frequency attack. In the following, we will describe each of them in detail.

\subsection{Patch-adaption Perturbation}
The ViTs divide the image into a group of non-overlapping patches and encode them into vectors with the same dimension~\citep{dosovitskiy2020image},~\citep{liu2021swin}, which makes ViTs highly sensitive to the modification of individual patch~\citep{fu2022patch}. 
Therefore, our attack focuses on modifications to the patches. Different from the existing low-frequency attack methods~\citep{guo2020low},~\citep{wang2022decision},~\citep{Li_2020_CVPR} that perform the DCT on the whole image, we first crop the image into a set of $16 \times 16 \times 3$ non-overlapping patches, and then perform a block DCT on all the patches separately. In doing so, the resultant DCT matrix has a patch attribute. This strategy plays an important role in breaking the semantics of the patch units.

For an input image $x_0$, we first crop out $n$ patches $P=[p_1, ..., p_n]^T\in R^{d\times d\times n}$ ($d$ is the width of a patch). Then, we perform the DCT on the $i^{th}$ patch:
\begin{equation}
        \label{eq2}
	\begin{split}
            F_i(u,v)=\frac{2}{d}\sum_{s=0}^{d-1}\sum_{t=0}^{d-1}(P_i)_{s,t}\textrm{cos}\frac{(2s+1)u\pi}{2d}\textrm{cos}\frac{(2t+1)v\pi}{2d},
	\end{split}	
\end{equation}
After transformation, $F_i(u,v)$ corresponds to the magnitude of wave
$\textrm{cos}\frac{(2u+1)v\pi}{2d}$, and lower values of $(u,v)$ present lower frequency. By performing the DCT for all patches of the image, we can get the DCT matrix of the entire image as
\begin{equation}
        \label{eq3}
	\begin{split}
            F=\bigcup_{i=1}^{n}F_i.
	\end{split}	
\end{equation}
The resultant matrix $F$ has a patch attribute, where the main content information is concentrated in the upper-left corner region of each DCT block.

\subsection{Weight Mask Matrix}
On one hand, most semantic information in natural images is located in the low-frequency part. Therefore, changes in the low-frequency components makes the model misclassify more easily. Based on this fact,
our method aims to generate low-frequency perturbations. We extract the low-frequency components of all the patches by multiplying the DCT matrix with a $0/1$ mask matrix. The $0/1$ mask matrix is also divided into different patches, and only $r \times r$ ($1\leq r\leq d$) positions in the upper-left corner of each patch have a value $1$, with the remaining positions being $0$. For a single patch $P_i \in R^{d \times d}$ of the image, the corresponding $0/1$ mask value is:
\begin{equation}
    1_{P_i}(s,t)=\left\{
                \begin{array}{ll}
                  1, \,\,\, 1\leq s,t\leq r,\\
                  0, \,\,\, others,
                \end{array}
              \right.
\end{equation}
where $r$ is a hyper-parameter that can control the number of low-frequency components of each patch to be attacked. Hence, we get the dimension reduction ratio:
\begin{equation}
    \rho = \frac{r}{d}.
\end{equation}

On the other hand, the human visual system is less sensitive to changes in the high-frequency regions of the image compared to the low-frequency regions. 
Therefore, the adversarial examples will have a better visual hiding effect when the perturbation focuses on the image's high-frequency areas. 
Since variance can reflect the degree of data fluctuation, we calculate the pixel variance of each patch and use it to represent the fluctuation level of the patch. For the patch sequence $P=[p_1,...,p_n]^T$ of the image, the corresponding variance value sequence $Q=[q_1,...,q_n]^T$ can be obtained. Considering the significant difference in the variance values among different patches, we calculate the normalized variance sequence $Q'=[q_1^{'},...,q_n^{'}]^T$ by performing the maximum normalization operation: 
\begin{equation}
    \label{eq6}
    q'_i = \frac{q_i}{\textrm{max}Q}.
\end{equation}
\noindent
Based on these, we can get the weight mask matrix:
\begin{equation}
        \label{eq7}
	\begin{split}
            M=\alpha \cdot \bigcup_{i=1}^{n}q'_i \cdot 1_{P_i},
	\end{split}	
\end{equation}
where the scaling factor $\alpha$ is a hyper-parameter,
which is used to avoid the situation where the values of the weight mask matrix are too small in most of the corresponding positions of the patches when the variance values are normalized. This is because an initial noise focusing on a few patch regions with a big weight value will result in the attack's failure.
The weight mask matrix can not only be used to limit the frequency components of the perturbation but can also highlight the importance of different patches by adjusting the proportion of distortion on each patch.

\begin{algorithm2e}[ht]
	\caption{Attack in Low-frequency DCT Domain}
	\label{alg1}
	\begin{algorithmic}[1]
		\REQUIRE Image $x_0$ and ground truth $y_0$, hard-label model $f(\cdot)$, max query number $N$, weight mask matrix $M$, $g = \textrm{inf}$.
		\ENSURE  Adversarial example $\hat{x}$
  		\FOR{$t=1,2,...,T$}
    		\STATE Randomly sample a perturbation $\theta$ from a Gaussian distribution	
                \STATE Get low-frequency perturbation $\theta'$ $\leftarrow$ $\theta\odot M$
                \STATE Compute $g(\theta')$ $\leftarrow$ Equation~\ref{eq10}
                \IF{$g(\theta') < g$}
    		      \STATE Update $g=g(\theta')$
                    \STATE Update $\theta'$ as the optimal initial perturbation
    		\ENDIF                 
            \ENDFOR 
		\WHILE{Query number $< N$}
			\STATE Randomly sample $\mu_1,...,\mu_J$ from a Gaussian or Uniform distribution
			\STATE Compute $\hat{\theta}$ $\leftarrow$ Equations~\ref{eq11} and~\ref{eq12}
			\STATE Update $\theta'$ $\leftarrow$ $\theta'-\eta\hat{\theta}$   
		\ENDWHILE 
		\STATE Get Adversarial example $\hat{x} \leftarrow$ Equation~\ref{eq13}
		\STATE \Return Adversarial example $\hat{x}$	
	\end{algorithmic}
\end{algorithm2e}

\subsection{Low-frequency Attack}
After obtaining the DCT matrix $F$ and the weight mask matrix $M$ of the original image, we adopt the Sign-OPT~\citep{ChengSCC0H20} to conduct adversarial attacks. However, unlike directly attacking in the pixel domain, our attack is performed in the DCT domain, and the frequency of being attacked is limited to a lower range, yielding a dimension reduction attack. Moreover, we design a weight mask matrix to control the proportion of patch perturbation. Finally, the inverse DCT (IDCT) can obtain the final adversarial example. The specific process can be divided into three steps.

\textbf{Step 1:} We generate a perturbation vector $\theta$ in the DCT domain randomly and then multiply it with the weight mask matrix $M$ to obtain a low-frequency perturbation vector:
\begin{equation}
	\begin{split}
            \theta'=\theta \odot M,
	\end{split}	
\end{equation}
where $\odot$ is the Hadamard product operation. Then, we calculate the shortest distance $g(\theta')$ required for obtaining an adversarial example in the direction $\frac{\theta'}{\parallel \theta' \parallel}$. 
We select the $\theta'$ which has the minimum $g(\theta')$ as the best initial perturbation vector on the DCT domain, and $g(\theta')$ as the initial distance $\lambda$, from a set of $\theta$. The calculation of a reasonable initial $\theta'$ is as follows:
\begin{equation}
       \label{eq9}
	\begin{split}
            \underset{\theta'}{\textrm{arg min}}\, g(\theta'),
	\end{split}	
\end{equation}
where 
\begin{equation}
       \label{eq10}
	\begin{split}
            g(\theta') = \underset{\lambda > 0}{\textrm{arg min}}\, (f(\textrm{IDCT}(F+\lambda\frac{\theta'}{\parallel \theta' \parallel})) \neq y_0),
	\end{split}	
\end{equation}
and $\lambda$ is computed through a binary search algorithm.

\begin{algorithm2e}[tp]
	\caption{Query-Efficient Hard-label Black-box Attack against Vision Transformers}
	\label{alg2}
	\begin{algorithmic}[1]
		\REQUIRE Image $x_0$ and ground truth $y_0$, hard-label model $f(\cdot)$, max query number $N$, scaling factor $\alpha$.
        \ENSURE  Adversarial example $\hat{x}$
        
        \STATE Crop image as $P=[p_1,...,p_n]^T$, 
        \STATE Compute DCT matrix $F$ $\leftarrow$ Equations~ \ref{eq2} and~\ref{eq3}
        \STATE Compute patch variance sequence $Q=[q_1,...,q_n]^T$
        \STATE Normalize $Q$ to $Q^{'}$ $\leftarrow$Equation~\ref{eq6}, 
        \STATE  Compute weight mask matrix 
        $M=\alpha \cdot \bigcup_{i=1}^{n}q'_i \cdot 1_{P_i}$
        \WHILE{Query number $< N$} 
            \STATE Search adversarial example $\hat{x}$ $\leftarrow$ Algorithm~\ref{alg1}
        \ENDWHILE
        \STATE \Return Reversed adversarial example $\hat{x}$ 
	\end{algorithmic}

\end{algorithm2e}

\textbf{Step 2:} When the initial $\theta'$ and the corresponding distance $g(\theta')$ are obtained, we need to determine further the update direction $\hat{\theta}$ of $\theta'$, and $\hat{\theta}$ needs to meet the condition that current adversarial example in the new direction $\theta'-\eta \hat{\theta}$  satisfies $g(\theta'-\eta \hat{\theta})<g(\theta')$, where $\eta$ is a step size factor. We use a sign gradient estimate method to calculate $\hat{\theta}$:
\begin{equation}
       \label{eq11}
	\begin{split}
            \hat{\theta} =\frac{1}{J}\sum_{j=1}^{J}\textrm{sign}(g(\theta'+\epsilon\mu_j)-g(\theta'))\mu_j\odot M,
	\end{split}	
\end{equation}
where $\mu_j$ is an randomly sampled Gaussian noise, and $J$ represents the number of random samples. $\textrm{sign}(\cdot)$ is the sign function which can be written as:
\begin{equation}
    \label{eq12}
    \textrm{sign}(g(\theta'+\epsilon\mu_j)-g(\theta')) = \left\{
                \begin{array}{ll}
                  +1, \,\,\, \textrm{if} \, f(x') == y_0,\\
                  -1, \,\,\, others,
                \end{array}
              \right.
\end{equation}
where
$x'=\textrm{IDCT}(F+g(\theta')\frac{\theta'+\epsilon\mu_j\odot M}{\parallel \theta'+\epsilon\mu_j\odot M \parallel})$.

\textbf{Step 3:} We can get the optimized $\theta'$ and $g(\theta')$ by repeating \textbf{Step 2}, within the maximum allowed number of model queries. The final adversarial example $\hat{x}$ can be obtained through the IDCT operation:
\begin{equation}
       \label{eq13}
    \hat{x}=\textrm{IDCT}(F+g(\theta')\frac{\theta'}{\parallel \theta'\parallel}).
\end{equation}

\begin{figure*}[tp]
	\centering	 
	\subfloat[]{
		\begin{minipage}{0.28\linewidth}   
			\centering	
			\includegraphics[width=1\textwidth]{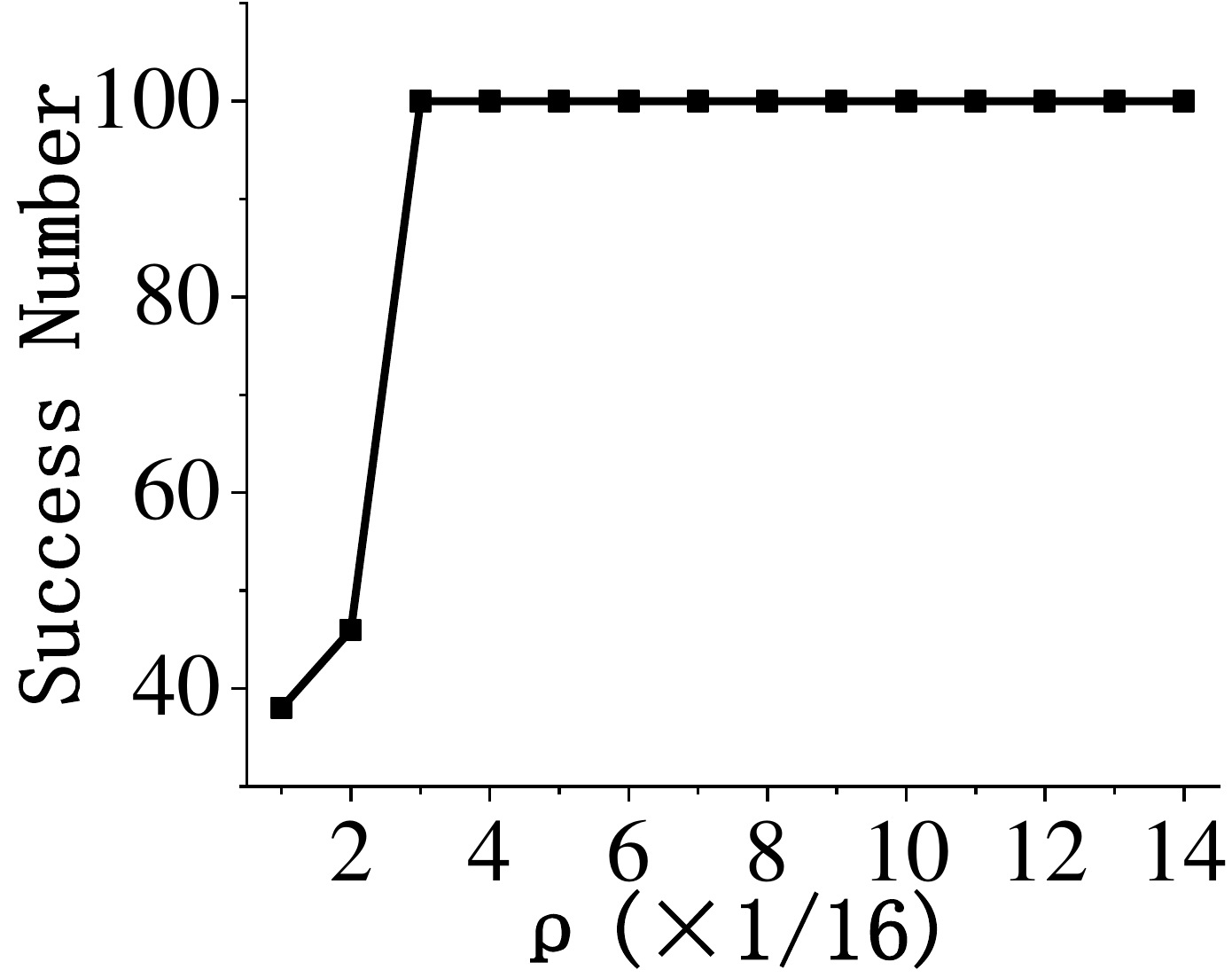}

		\end{minipage}}
  \quad 
	\subfloat[]{
		\begin{minipage}{0.28\linewidth}
			\centering
			\includegraphics[width=1\textwidth]{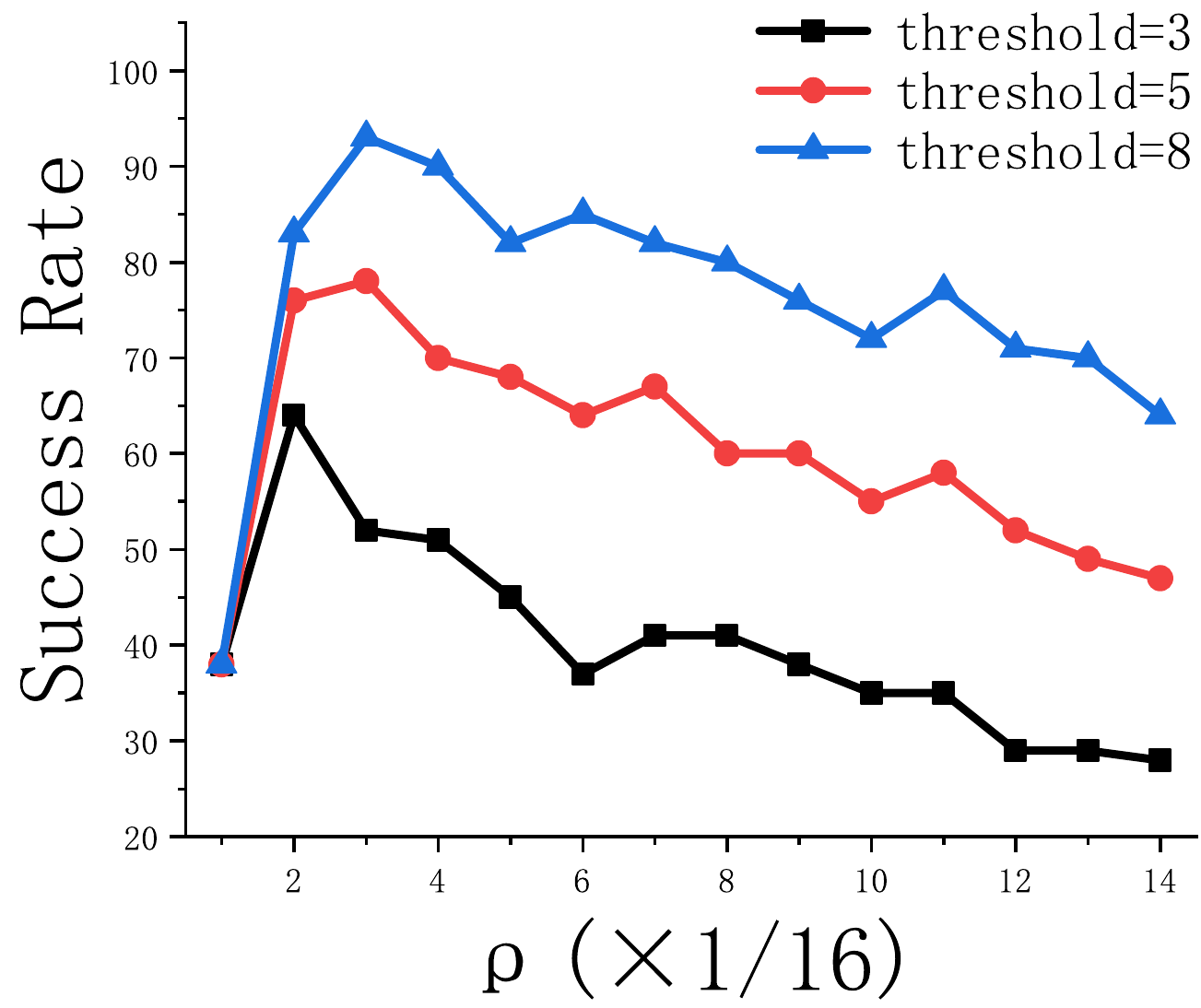}
		\end{minipage}
	}
 \quad 
	\subfloat[]{
	\begin{minipage}{0.28\linewidth}
		\centering
		\includegraphics[width=1\textwidth]{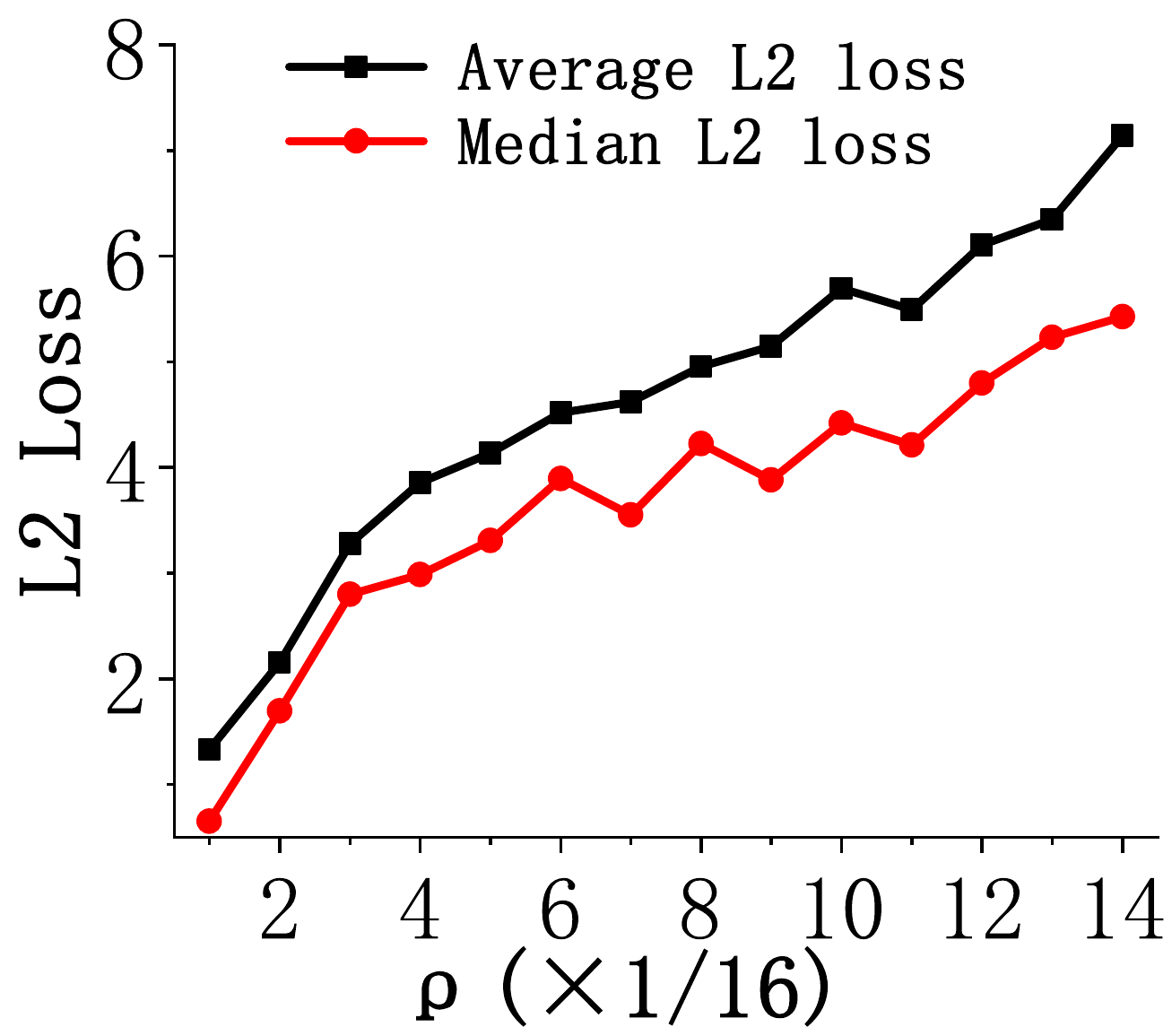}
	\end{minipage}
}
\quad
	\caption{Results of AdvViT+ with different $\rho$ values on the Deit-T model. Within the maximum number of model queries 4000: (a) is the number of samples that can successfully find adversarial examples in 100 tested images; (b) shows the success rates on the given thresholds ($\epsilon$) 3.0, 5.0 and 8.0, respectively; (c) gives out the $L_2$ loss curves, including average $L_2$ loss and median $L_2$ loss.}
  	\label{fig3}
	\vspace{-0.5cm}	
\end{figure*}

Algorithm~\ref{alg1} describes the detailed attack process in the low-frequency DCT domain and generates the final adversarial examples. Through the above process, we can clearly observe that the adversarial examples generated by our method are built patch-wise with low-frequency perturbation on each patch. Therefore, our attack actualizes query efficiency against the ViTs that process images in a patch-wise manner. Algorithm~\ref{alg2} shows the overall procedure of our proposed AdvViT which takes the Sign-OPT attack~\citep{ChengSCC0H20} as its baseline. 
Note that our method can be easily integrated with other baselines to achieve effective attacks on ViTs. For example, in this article, we also utilize the Sign-OPT+ attack~\citep{ran2022sign}, which has shown a performance improvement compared to the Sign-OPT attack, as a baseline to conceive a new attack method named AdvViT+.

\begin{table}[tp]
	\centering
         \setlength{\tabcolsep}{17pt}
        \renewcommand{\arraystretch}{1.1}
   \caption{Performance of AdvViT+ with different patch sizes on the Deit-T model, and the $\rho$ is set to $3/16$ uniformly. The results indicate that AdvViT+ achieves the best attack effect when employing a patch size of $16\times16$, which attains the smallest distortion loss. The best competitor is highlighted in bold.}
    	\begin{tabular}{c|cc}
    		\hline
    		Patch size &Average $L_2$ loss&Median $L_2$ loss\\ 
    		\hline			
            $8\times8$  & 3.6526   & 3.0522    \\		
            $16\times16$&  \textbf{3.1761} $\downarrow$ &  \textbf{2.4715} $\downarrow$  \\		
    	$20\times20$&  3.7016  &   3.7802  \\	
     
    		\hline  
    	\end{tabular}

\label{tab6}
\end{table}

\section{Experimental Results}
\subsection{Datasets and Target Models}
We evaluate our algorithm on the validation set of ImageNet-1K~\citep{deng2009imagenet}.
For fair comparison, eight classification models are used for testing, including ResNet-50~\citep{he2016deep}, VGG-16~\citep{vgg2015m}, Deit models~\citep{touvron2021training} (including Deit-T, Deit-S, Deit-B), and Swin Transformers~\citep{liu2021swin} (including Swin-T, Swin-S, Swin-B). Our AdvViT and AdvViT+ are compared with six leading hard-label black-box attack methods, which are Opt attack~\citep{cheng2019query}, Sign-OPT attack and SurFree attack~\citep{Maho_2021_Sur}, Sign-OPT+ attack~\citep{ran2022sign}, Sign-OPT DCT attack and the PAR attack~\citep{shi2022decision}. The Sign-OPT DCT attack is another version of the Sign-OPT attack, and it only changes the attack domain from the pixel domain to the DCT domain. The AdvViT+ attack is the method we proposed using the Sign-OPT+ as the baseline. For convenience, we use abbreviations in the following tables and figures as Sign: Sign-OPT attack, SD: Sign-OPT DCT attack, Sign+: Sign-OPT+ attack, AD: AdvViT attack, AD+: AdvViT+ attack.
\begin{figure*}[tp]
	\centering	 
	\subfloat[]{
		\begin{minipage}{0.28\linewidth}   
			\centering	
			\includegraphics[width=1\textwidth]{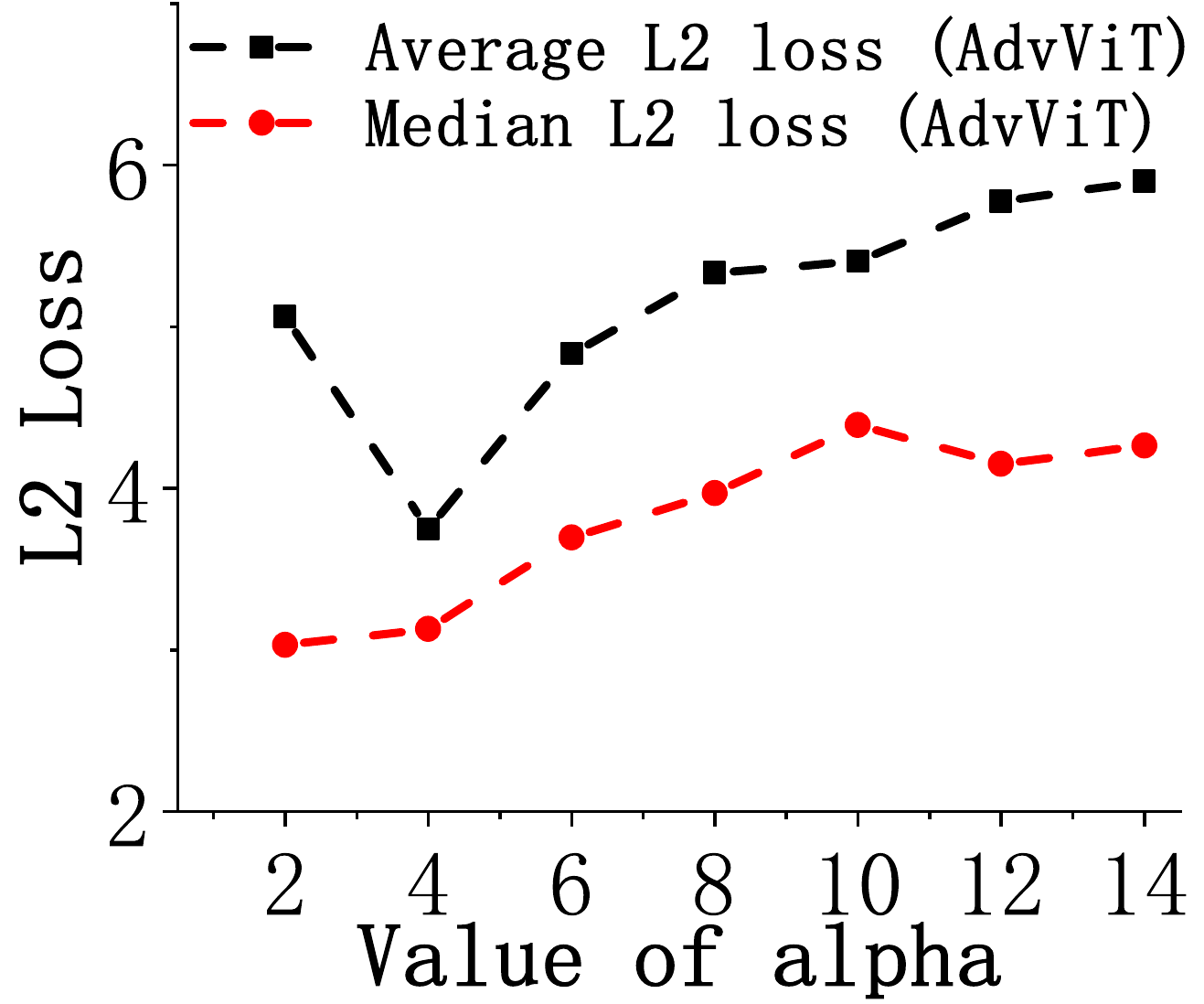}

		\end{minipage}}
  \quad 
	\subfloat[]{
		\begin{minipage}{0.28\linewidth}
			\centering
			\includegraphics[width=1\textwidth]{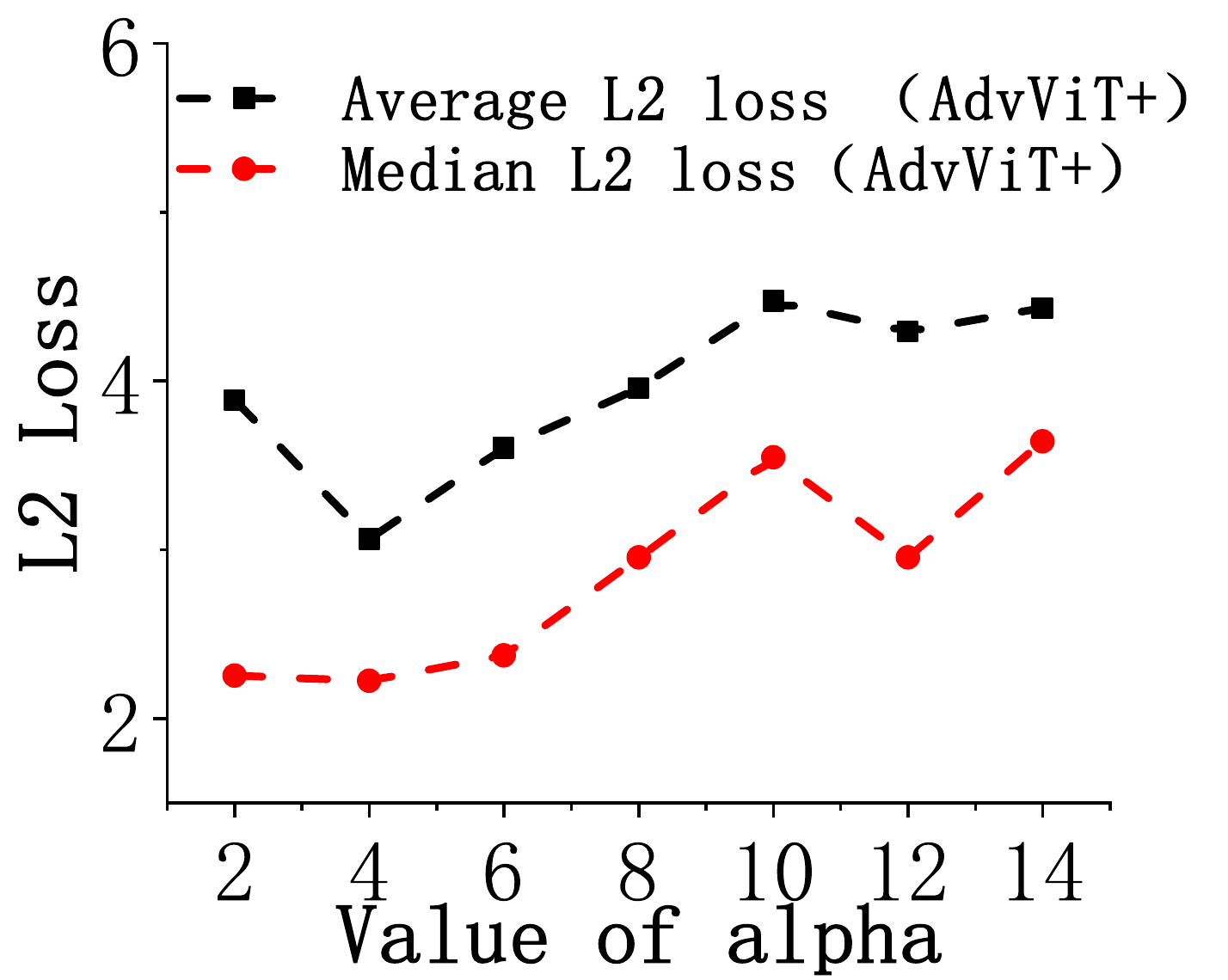}
		\end{minipage}
	}
 \quad 
	\subfloat[]{
	\begin{minipage}{0.28\linewidth}
		\centering
		\includegraphics[width=1\textwidth]{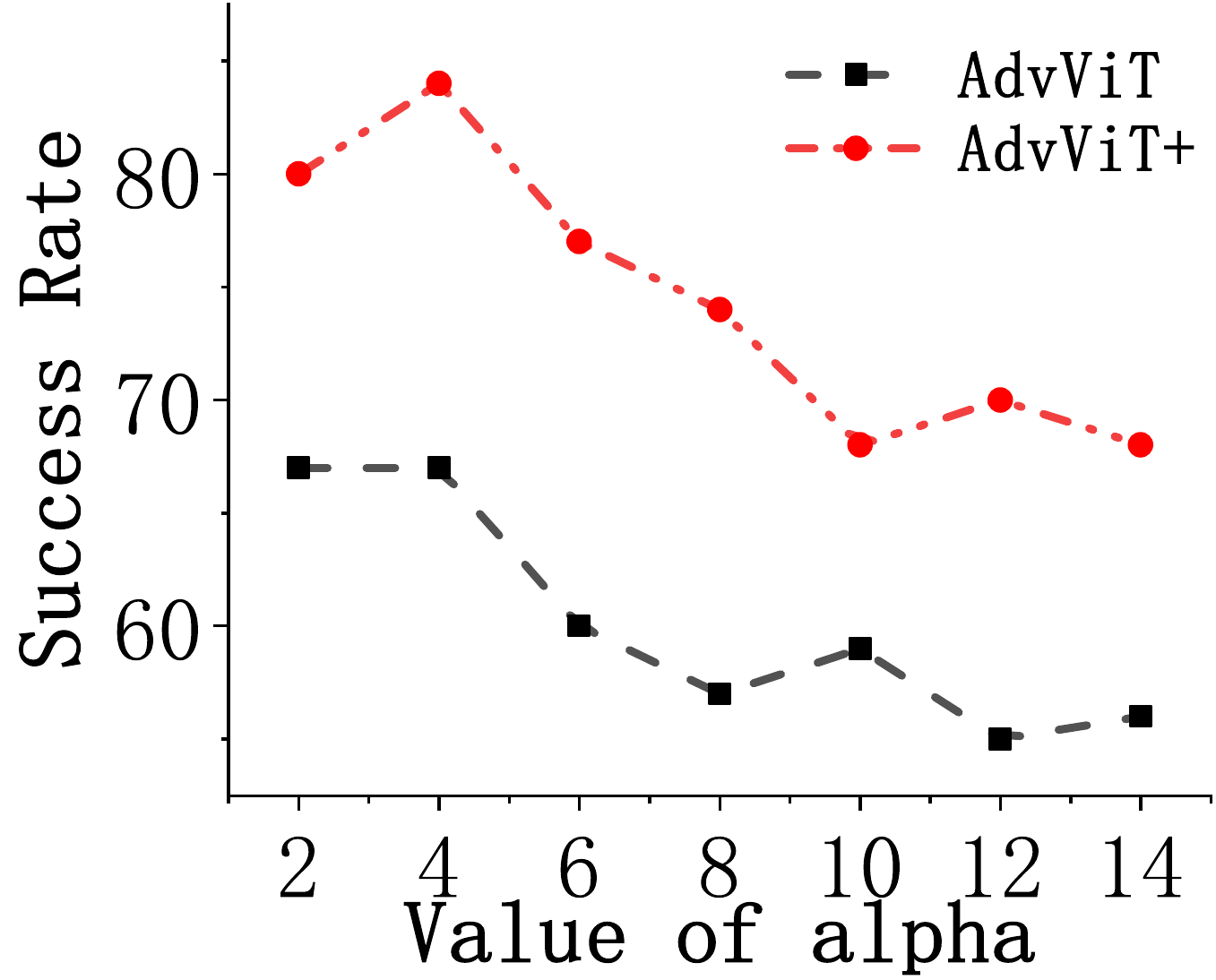}
	\end{minipage}
}
\quad
	\caption{Results of AdvViT and AdvViT+ with different $\alpha$ values on the Deit-T model. The maximum number of model queries is 4000 and the dimension reduction ratio $\rho=3/16$. (a) and (b) show the $L_2$ loss curves of AdvViT and AdvViT+, respectively. (c) shows the success rate curves on the given threshold $\epsilon =5.0$.}
  	\label{fig4}
	\vspace{-0.5cm}	
\end{figure*}


\subsection{Metrics and Parameters Selection}
For fair comparison, we randomly select 100 original images from the validation set, and generate adversarial examples for each image. 
Four important metrics are selected to evaluate the black-box adversarial attacks, namely $L_2$ loss (including average loss and median loss), success rate, PSNR, and SSIM. 
The success rate is the percentage of images that can obtain adversarial examples with distortion below the given threshold $\epsilon$ within the maximum number of model queries. 
In this paper, we empirically set $\epsilon$ to $5.0$.

\begin{table*}[tp]
	\centering
\caption{Performance comparison between various methods in the $L_2$-norm setting. The best competitor is highlighted in bold.}
 	\fontsize{10}{10}\selectfont
	\setlength{\tabcolsep}{5mm} 
	{
         \setlength{\tabcolsep}{2pt}
        \renewcommand{\arraystretch}{1.4}
 
	\begin{tabular}{cc|cccccccc}
		\hline
            &Models&Opt&SurFree&Sign&SD&PAR&AD&Sign+&\textbf{AD+}\\
		\hline				
        \multirow{8}{*}{\makecell{Average\\ Loss}}&ResNet50&12.8751&8.0953&4.2523&13.2774&6.3018&6.2893&\textbf{4.8720}&6.3144\\	&VGG16&9.3259&14.7941&\textbf{2.0789}&7.2465&3.6063&4.4646&3.0132&4.0384\\
  		\cline{2-10} 
	   &Deit-S&16.7671&5.4658&15.5463&17.5765&31.5332&7.4404&6.6530&\textbf{4.0624}\\		
		&Deit-T&11.4361&6.5564&11.7626&19.5180&13.7337&3.7447&3.9968&\textbf{3.0631}\\
		&Deit-B&17.8781&6.2534&12.6837&22.5207&52.4684&8.1070&6.4474&\textbf{5.7619}\\		
		&Swin-S&24.3537&10.3771&18.1489&23.3706&48.3882&12.1214&9.7330&\textbf{7.2964}\\		
		&Swin-T&18.9859&9.4796&13.0315&17.6031&38.8738&8.4900&6.3743&\textbf{4.7670}\\	
		&Swin-B&22.5757&8.0298&17.4582&24.7072&47.3906&15.3613&11.2282&\textbf{6.8812}\\	
		\hline
            \multirow{8}{*}{\makecell{Median \\Loss}}&ResNet50&11.5817&2.5227&\textbf{2.4818}&10.1293&4.5236&5.0530&3.3934&4.4145\\	&VGG16&8.0083&1.7939&\textbf{0.4336}&6.3378&1.4762&2.9580&2.4211&3.2532\\
  		\cline{2-10} 
	   &Deit-S&16.4029&3.8233&15.5697&14.8415&19.7330&5.3299&5.6825&\textbf{3.3766}\\		
		&Deit-T&9.9202&2.8067&11.2101&16.4300&10.5683&3.1298&3.2085&\textbf{2.2241}\\
		&Deit-B&17.5661&3.9820&10.5546&18.8709&19.8792&5.2914&5.2341&\textbf{3.8227}\\		
		&Swin-S&23.7568&\textbf{4.7645}&16.3866&16.6014&29.3052&8.5019&7.3851&6.3456\\		
		&Swin-T&17.0676&5.2852&10.2822&13.6054&19.5116&5.8202&4.4812&\textbf{3.6390}\\	
		&Swin-B&23.4078&\textbf{5.1101}&13.2832&19.7776&23.3974&8.6092&8.8189&5.5994\\	
		\hline
	\end{tabular} 
 \vspace{-0.7cm}	
\label{tab1}
}
\end{table*}

On one hand, the dimension reduction ratio $\rho$ controls the percentage of low-frequency components of each patch to be attacked. A too-small $\rho$ will result in a failure attack. We test $14$ values of $\rho$ by taking $r$ within $[1, 14]$ with a unit interval, and then calculate the $L_2$ loss and success rate. 
Figure~\ref{fig3} shows the results of AdvViT+ with different $\rho$ values on the Deit-T model. 
It can be seen from Figure~\ref{fig3} (a) that the attack cannot successfully get adversarial examples for every original image when the $\rho$ is set to too small, such as $1/16$ and $2/16$. Nevertheless, Figure~\ref{fig3} (b-c) shows that a smaller $\rho$ will result in a higher success rate and a lower $L_2$ loss. Therefore, to ensure that all original images can be transformed into adversarial examples, we take $\rho$ as small as possible, \textit{i.e.} $\rho=3/16$.

Based on the best value $\rho=3/16$, we test the performance of AdvViT+ with different patch sizes on the Deit-T model, including $8\times8$, $16\times16$, $20\times20$. The results indicate that AdvViT+ with patch size $16\times16$ achieves the best attack effects, and the smallest average and median adversarial losses. Taking it a step forward, we can conclude that the optimal attack effect is attained when the patch size of the attack units exactly matches that of the data unit within ViT. That is $16\times16$. 

On the other hand, the hyper-parameter $\alpha$ is used to avoid the situation where the normalized variance values are too small, which can lead to a failure in finding initial noise. However, an excessive $\alpha$ also leads to performance degradation. 
We take seven values for $\alpha$, which are within $[2, 14]$ with two unit intervals. 
For each value, we simultaneously record the results of AdvViT and AdvViT+ on the Deit-T model and the dimension reduction ratio $\rho=3/16$. 
All the results are plotted as curves in Figure~\ref{fig4}. Figure~\ref{fig4} (a) and (b) show the $L_2$ loss curves of AdvViT and AdvViT+, respectively, and Figure~\ref{fig4} (c) gives the success rate curves on the given threshold $\epsilon =5.0$.
It can be seen from Figure~\ref{fig4} that when the value of $\alpha$ is $4$, both the average $L_2$ loss and median $L_2$ loss of the two proposed methods are relatively lower, while the success rates are highest. Based on this observation, we take a value of $\alpha =4$.

\begin{table}[tp]
	\centering
          \setlength{\tabcolsep}{7pt}
   \caption{Comparison of attack success rates of various methods in the $L_2$-norm setting. 
   }
    	  	\fontsize{10}{10}\selectfont
	\setlength{\tabcolsep}{15mm} 
	{
         \setlength{\tabcolsep}{8pt}
        \renewcommand{\arraystretch}{1.4}
        
        \begin{tabular}{c|cccccccc}
    		\hline
    		Models &Opt&SurFree&Sign&SD&PAR&AD&Sign+&\textbf{AD+} \\ 
    		\hline			
    		ResNet50&25\%&\textbf{70\%}&67\%&26\%&53\%&50\%&67\%&58\%\\		
    		VGG16&34\%&74\%&\textbf{85\%}&42\%&74\%&66\%&82\%&75\%\\		
                \hline
    		Deit-S&13\%&60\%&11\%&13\%&17\%&47\%&45\%&\textbf{72\%}\\		
    		Deit-T&24\%&68\%&24\%&18\%&33\%&67\%&68\%&\textbf{84\%}\\
    		Deit-B&13\%&59\%&21\%&16\%&23\%&48\%&46\%&\textbf{60\%}\\		
    		Swin-S&9.0\%&\textbf{51\%}&16\%&15\%&15\%&33\%&34\%&34\%\\		
    		Swin-T&16\%&47\%&22\%&24\%&14\%&44\%&53\%&\textbf{70\%}\\	
    		Swin-B&13\%&50\%&18\%&13\%&17\%&34\%&31\%&\textbf{50\%}\\  
    		\hline  
    	\end{tabular}
\label{tab2}}
\end{table}

\subsection{Attack Results}
Table~\ref{tab1} shows the adversarial losses of different attack methods on the validation set of ImageNet-1K, regarding both average $L_2$ loss and median $L_2$ loss separately. Table~\ref{tab1} reveals that our method does not outperform the CNN-based ones like ResNet50 and VGG16. In these cases, the average loss is not as good as the Sign-OPT+ attack (on ResNet50) and Sign-OPT attack (on VGG16), and the median loss is higher than the Sign-OPT attack. Instead, our attack demonstrates excellent performance in targeting ViT models. Firstly, the average losses of the adversarial examples generated by our method, AdvViT+, are significantly lower than the state-of-the-art methods on all the ViT models. For the median losses, the AdvViT+ attack performs better on the Deit-S, Deit-T, Deit-B, and Swin-T models.
Secondly, our AdvViT and AdvViT+ methods achieve significant improvements compared to their baselines Sign-OPT attack and Sign-OPT+ attack, respectively. Compared with the Sign-OPT DCT attack, which also performs attacks in the DCT domain,  AdvViT achieves a much better attack effect. This study provides further validation of the effectiveness of the proposed attack strategy, emphasizing that targeting the low-frequency components of each patch unit enhances attack efficiency. The performance of the AdvViT attack is not the best due to the constraint on the baseline. Promisingly, our attack strategy can be easily combined with the existing attack methods to achieve a better performance, such as our AdvViT+, which takes the Sign-OPT+ attack as the baseline. In terms of the attack success rate, Table~\ref{tab2} provides the results of the corresponding attacks in Table~\ref{tab1}. The results show that our AdvViT+ method achieves the highest attack success rate on all ViT models except for Swin-S. Furthermore, the success rates of the proposed methods, AdvViT and AdvViT+, show significant improvements over their corresponding baselines for the tested ViT models.

In terms of the robustness comparison between CNNs and ViTs, it is not difficult to discover the following facts from Tables~\ref{tab1} and~\ref{tab2}: 
i) When tested with the attack methods specifically designed for CNN-based models, such as OPT, SurFree, SD, Sign-OPT, and Sign-OPT+ attacks, ResNet50, and VGG16, our method exhibits inferior robustness compared to ViT models. These CNN models demonstrate lower average and median losses, as well as higher attack success rates.
ii) When testing on our AdvViT+, the robustness of CNN-based models exhibits different results. For instance, the attack success rate of ResNet50 is lower than that of Deit-T, Deit-S, Deit-B, and Swin-T, while the attack success rate of VGG16 is also lower than that of Deit-T. This pattern also holds for the adversarial losses. iii) After incorporating our attack strategies, the proposed AdvViT and AdvViT+ methods show significant performance improvements in attacking the ViT-based models compared to their respective baseline Sign-OPT and Sign-OPT+ attacks. However, when applying these strategies to CNN-based models, the opposite scenario is observed.
\begin{table*}[tp]
	\centering
   \caption{Comparison between various methods of the PSNR and SSIM on various models. 
   }
 	\fontsize{10}{10}\selectfont
	\setlength{\tabcolsep}{5mm} 
	{
         \setlength{\tabcolsep}{2pt}
        \renewcommand{\arraystretch}{1.4}

	\begin{tabular}{cc|cccccccc}
		\hline
		
            &Models&Opt&SurFree&Sign&SD&PAR&AD&Sign+&\textbf{AD+} \\
		\hline
        \multirow{8}{*}{SSIM}&ResNet50&0.9239&0.9845&0.9833&0.8965&0.9663&0.9825&0.9809&\textbf{0.9852}\\	 
        &VGG16&0.9421&0.9881&\textbf{0.9940}&0.9521&0.9613&0.9891&0.9881&0.9891\\	
            \cline{2-10}  
		&Deit-S&0.8942&0.9752&0.8617&0.8365&0.9188&0.9849&0.9656&\textbf{0.9858}\\		
		&Deit-T&0.9291&0.9807&0.8958&0.9262&0.9498&0.9895&0.9780&\textbf{0.9908}\\
		&Deit-B&0.8888&0.9759&0.9033&0.7447&0.9196&0.9800&0.9673&\textbf{0.9868}\\		
		&Swin-S&0.8410&0.9657&0.8384&0.7860&0.8716&0.9616&0.9418&\textbf{0.9718}\\		
		&Swin-T&0.9009&0.9638&0.9146&0.8579&0.9170&0.9745&0.9704&\textbf{0.9842}\\	
		&Swin-B&0.8514&0.9648&0.8596&0.7487&0.8740&0.9619&0.9370&\textbf{0.9710}\\		
		\hline
        \multirow{8}{*}{PSNR}&ResNet50&27.7817&\textbf{40.2737}&35.1089&32.9933&34.5341&38.3017&37.1755&39.4922\\	 
        &VGG16&30.3642&\textbf{41.5478}&37.4577&36.1942&33.3788&40.3389&38.5173&40.5504\\	
            \cline{2-10}  
		&Deit-S&25.3887&39.0821&25.2505&29.6551&27.7183&38.2919&35.1772&\textbf{40.5196}\\		
		&Deit-T&28.2624&40.5158&27.1805&34.0795&31.2891&40.8737&37.3764&\textbf{41.6160}\\
		&Deit-B&24.6986&39.0938&27.0031&27.3333&27.6262&37.5120&35.3763&\textbf{39.8926}\\		
		&Swin-S&23.0587&\textbf{37.1734}&25.6615&28.4512&24.5452&34.9810&33.3260&36.6700\\		
		&Swin-T&25.7812&38.0812&28.8429&31.0247&27.8844&37.3245&38.0812&\textbf{39.1802}\\	
		&Swin-B&22.9373&\textbf{38.1958}&25.3890&27.2089&23.8724&34.4899&33.0630&37.3000\\		
		\hline
	\end{tabular} 
 	\label{tab3}
  \vspace{-0.5cm}	}
\end{table*}

Based on the above observations, we can draw a clear conclusion: ViTs are not always more robust than CNNs in hard-label black-box scenarios. The earlier viewpoint of ViTs being more robust than CNNs in some studies might be attributed to the fact that the evaluation methods they used were originally designed for CNNs. We found that these methods did not fully consider the unique characteristics of ViTs, namely their sensitivity to modifications at the patch unit level. In contrast, our work considers this specific characteristic, leading to a more accurate assessment of the robustness of ViTs compared to the methodology built on CNNs.
    
\subsection{Visual Effect Results}
The visual effect is used to measure the degree to which perturbations in adversarial examples can be detected by the human eye.
To show the advantage of the proposed method on the visual effect, we conduct a performance comparison from the quantitative and qualitative perspectives.
\begin{figure*}[ht]
        \centering
    		\begin{minipage}{0.11\linewidth} 
    			    Original
 	\subfloat[-\\-]{
    			\centering	
    	\includegraphics[width=1\textwidth]{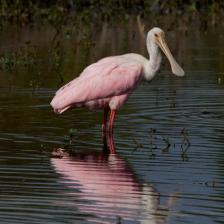}}
    		\end{minipage}    
    		\begin{minipage}{0.11\linewidth}
    		    	 Opt	
	\subfloat[0.8565,\\21.9895]{ 
    			\centering
      \includegraphics[width=1\textwidth]{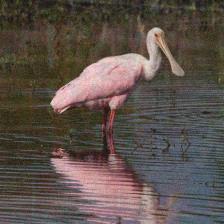}}
      		\end{minipage}
    		\begin{minipage}{0.11\linewidth}
    			Sign    			
 	\subfloat[0.5822,\\14.6600]{
 		    			\centering	
    \includegraphics[width=1\textwidth]{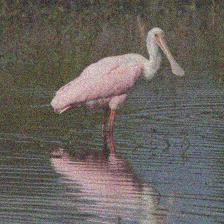}}
    		\end{minipage}
    	    \begin{minipage}{0.11\linewidth}  
    	    	    			PAR 
 	\subfloat[0.8506,\\22.2686]{
    			\centering	
    \includegraphics[width=1\textwidth]{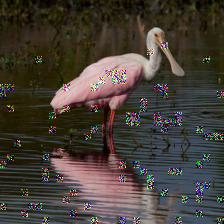}} 
    		\end{minipage}    
    	    \begin{minipage}{0.11\linewidth}
    	    	    Sign+ 
	\subfloat[0.8407,\\30.2621]{
    			\centering
    \includegraphics[width=1\textwidth]{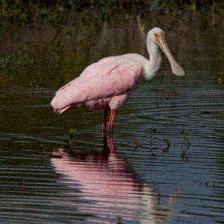}} 
    		\end{minipage}
    	    \begin{minipage}{0.11\linewidth}
    	    	   SurFree
	\subfloat[0.8661,\\30.9510]{
    			\centering
    \includegraphics[width=1\textwidth]{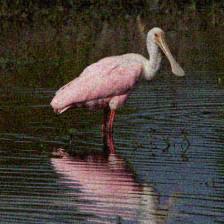}} 
    		\end{minipage}
    	    \begin{minipage}{0.11\linewidth}
    	    	    AdvViT
	\subfloat[\textbf{0.9884},\\40.1752]{
    			\centering
    \includegraphics[width=1\textwidth]{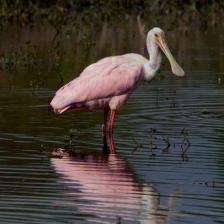}}
    		\end{minipage}
    	     \begin{minipage}{0.11\linewidth}
    	    	AdvViT+
     \subfloat[0.9870,\\ \textbf{40.1808}]{
        			\centering
    \includegraphics[width=1\textwidth]{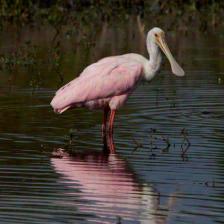}}    
    		\end{minipage}   
    \quad
    
        \begin{minipage}{0.11\linewidth}
        \subfloat[-\\-]{
     			\centering	
    		\includegraphics[width=1\textwidth]{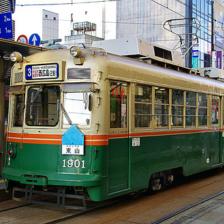}}
             \end{minipage}
         \begin{minipage}{0.11\linewidth}
	\subfloat[0.8375,\\20.2709]{
    			\centering
    			\includegraphics[width=1\textwidth]{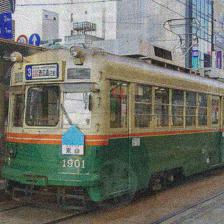}}
    		    		\end{minipage}
    	   \begin{minipage}{0.11\linewidth} 
 	\subfloat[0.8178,\\22.0349]{
    			\centering	
    			\includegraphics[width=1\textwidth]{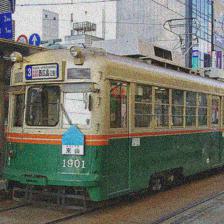}}    		\end{minipage}
    		 \begin{minipage}{0.11\linewidth} 
       \subfloat[0.8817,\\22.3979]{
    			\centering	
            \includegraphics[width=1\textwidth]{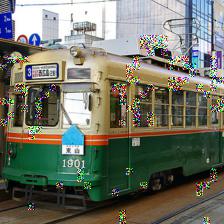}}
               \end{minipage}
     		\begin{minipage}{0.11\linewidth}          
	\subfloat[0.9414,\\32.9237]{
    			\centering
    			\includegraphics[width=1\textwidth]{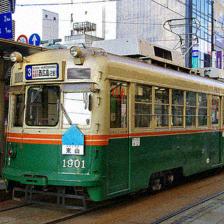}}     		
       \end{minipage}
    		\begin{minipage}{0.11\linewidth}
	\subfloat[0.8418,\\28.7262]{
    			\centering
    			\includegraphics[width=1\textwidth]{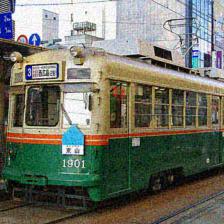}}     		\end{minipage}
    		\begin{minipage}{0.11\linewidth}
	\subfloat[0.9775,\\36.7450]{
    			\centering
    			\includegraphics[width=1\textwidth]{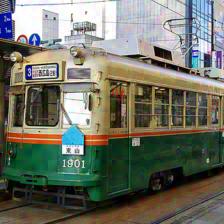}}    		
    		\end{minipage}
    	    \begin{minipage}{0.11\linewidth}
	\subfloat[\textbf{0.9898,}\\\textbf{40.4693}]{
    			\centering
    			\includegraphics[width=1\textwidth]{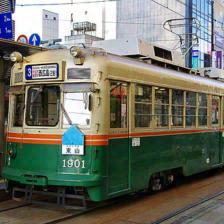}}
    		    		\end{minipage}       
     \quad
            		\begin{minipage}{0.11\linewidth}
        \subfloat[-\\-]{
       			\centering	
    		\includegraphics[width=1\textwidth]{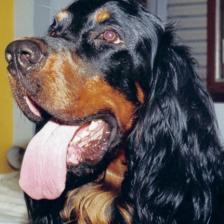}}
                		\end{minipage}
    		\begin{minipage}{0.11\linewidth}
	\subfloat[0.7181,\\18.4578]{
    			\centering
    			\includegraphics[width=1\textwidth]{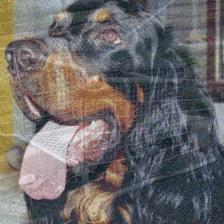}}
    		    		\end{minipage}
    		\begin{minipage}{0.11\linewidth}     	    		
 	\subfloat[0.6773,\\18.8537]{
    			\centering	
    			\includegraphics[width=1\textwidth]{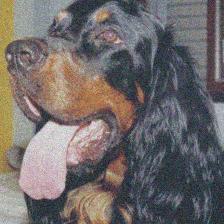}}
    		    		\end{minipage}
    	 \begin{minipage}{0.11\linewidth}  
       \subfloat[0.8612,\\24.7961]{
     			\centering	
            \includegraphics[width=1\textwidth]{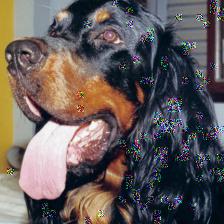}} 
        		\end{minipage}     
    		\begin{minipage}{0.11\linewidth}
	\subfloat[0.9233,\\32.9328]{
    			\centering
    			\includegraphics[width=1\textwidth]{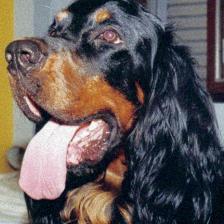}}     		
       \end{minipage}
    		\begin{minipage}{0.11\linewidth}    		
	\subfloat[0.9231,\\32.9834]{
    			\centering
    			\includegraphics[width=1\textwidth]{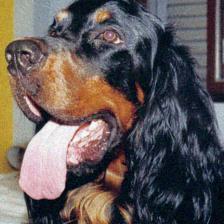}}           		\end{minipage}
    		\begin{minipage}{0.11\linewidth}
	\subfloat[0.9675,\\34.2842]{
    			\centering
    			\includegraphics[width=1\textwidth]{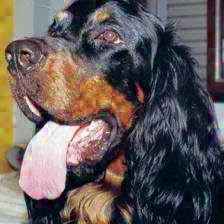}}          		\end{minipage}
    		\begin{minipage}{0.11\linewidth}
	\subfloat[\textbf{0.9733,}\\ \textbf{36.0095}]{
    			\centering
    			\includegraphics[width=1\textwidth]{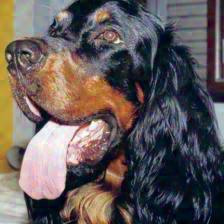}} 
    		    		\end{minipage}      
     \quad

 \caption{Comparison of visual effect against the Deit-T model. The first and second values in parentheses are SSIM and PSNR, respectively. We can easily observe that our methods have better visual effects. It is because the perturbations of our method are concentrated in the high-frequency regions where the gradient change is usually larger than that of the smooth regions, and the changes in these areas are not easily noticed by human eyes.}
 	\label{fig5}
\end{figure*}

From a quantitative perspective, we employ two renowned metrics, PSNR and SSIM, to gauge the similarity of the original images and adversarial examples. Larger values of these metrics indicate higher similarity, suggesting that the adversarial examples are less likely to be discerned by the human eye.
Table~\ref{tab3} compares the median PSNR and SSIM values. AdvViT+ attains the highest SSIM values across all tested classification models except for VGG-16, where it is surpassed by the Sign-OPT attack. Additionally, median PSNR results are generally higher in most ViT cases. Consequently, our attack demonstrates superior visual performance in targeting ViT models.

From a qualitative perspective, Figure~\ref{fig5} shows the adversarial examples generated by six different methods on three original images against the Deit-T model.
We can easily observe from Figure~\ref{fig5} that our methods have better visual effects. It is because the perturbations of our method are concentrated in the high-frequency regions where the gradient change is usually larger than that of the smooth regions, and the changes in these areas are not easily noticed by human eyes. However, the perturbations in the adversarial examples of PAR can be easily observed, which results from concentrating the perturbations to a few patches.

\subsection{Ablation Studies}

\begin{table*}[tp]
        \setlength{\tabcolsep}{3pt}
        \renewcommand{\arraystretch}{1.1}
	\centering
   \caption{The ablation study on the contribution of our innovative design towards the whole framework. Method \textbf{A} attacks all the components of the DCT domain, without dimension reduction. While method \textbf{B} utilizes patch-wise DCT and only attacks the low-frequency components, without using the weight mask matrix.}
  	\fontsize{10}{10}\selectfont
	\setlength{\tabcolsep}{15mm} 
	{
         \setlength{\tabcolsep}{6pt}
        \renewcommand{\arraystretch}{1.4}
        
    	\begin{tabular}{c|ccc|ccc|ccc}
    		\hline
    		\multirow{2}{*}{Models}&\multicolumn{3}{c|}{ASR (\%)} &\multicolumn{3}{c|}{SSIM}&\multicolumn{3}{c}{PSNR}\\
      		\cline{2-10}
                &A&B&AD+&A&B&AD+&A&B&AD+\\
    		\hline			
    		Deit-S&13&63&\textbf{72}&0.8365&0.9773&\textbf{0.9858}&29.6551&39.1472&\textbf{40.5196}\\		
    		Deit-T&18&81&\textbf{84}&0.9262&0.9900&\textbf{0.9908}&34.0795&41.5970&\textbf{41.6160}\\
    		Deit-B&16&60&\textbf{60}&0.7447&0.9812&\textbf{0.9868}&27.3333&39.2412&\textbf{39.8926}\\		
    		Swin-S&15&33&\textbf{34}&0.7860&0.9555&\textbf{0.9718}&28.4512&36.0560&\textbf{36.6700}\\		
    		Swin-T&24&52&\textbf{70}&0.8579&0.9725&\textbf{0.9842}&31.0247&38.5185&\textbf{39.1802}\\	
    		Swin-B&13&46&\textbf{50}&0.7487&0.9621&\textbf{0.9710}&27.2089&37.2114&\textbf{37.3000}\\ 
    		\hline  
    	\end{tabular}
\label{tab5}
	\vspace{-0.5cm}	}
\end{table*}

To clarify the effectiveness and contribution of our innovative design, we conduct two ablation experiments: \textbf{A}) Attacking all the components of the DCT domain without dimension reduction; \textbf{B}) Using the patch-wise DCT to attack only the low-frequency components without the weight mask matrix. As shown in Table \ref{tab5}, compared with \textbf{A} and \textbf{B}, our method achieves significant performance improvements in terms of ASR, SSIM, and PSNR. Note that the superiority of our AdvViT+ over \textbf{B} in all three metrics provides solid evidence for the effectiveness of the weight mask matrix.

\section{Conclusion}
In this paper, we take the first step towards exploring the intrinsic nature of ViTs to investigate their vulnerability, and have presented a query-efficient black-box attack against ViTs.
Our attack strategy is designed based on the following facts: 1) Patch-based ViTs are more sensitive to the modifications on patches; 2) It is effective in optimizing adversarial perturbation on the low-frequency domain of each patch, reflecting the semantic switching of a local small object; 3) The changes in the high-frequency region of the whole image are less noticeable to the human eyes. Experimental results have shown the efficiency and advantages of our method with the same query numbers. Our method further reveals the vulnerability of ViTs, which reverses previous views that ViTs are more robust than CNNs against adversarial attacks. This work is expected to shed new insight into the robustness of ViTs.

\bibliographystyle{unsrtnat} 
\bibliography{main}

\end{document}